\documentclass[sigconf,natbib=false]{acmart}
\AtBeginDocument{%
  }

\setcopyright{acmlicensed}
\copyrightyear{2026}
\acmYear{2026}
\setcopyright{cc}
\setcctype{by}
\acmConference[KDD '26]{Proceedings of the 32nd ACM SIGKDD Conference on Knowledge Discovery and Data Mining V.2}{August 09--13, 2026}{Jeju Island, Republic of Korea}
\acmBooktitle{Proceedings of the 32nd ACM SIGKDD Conference on Knowledge Discovery and Data Mining V.2 (KDD '26), August 09--13, 2026, Jeju Island, Republic of Korea}
\acmDOI{10.1145/3770855.3818954}
\acmISBN{979-8-4007-2259-2/2026/08}

\newcommand{\github}{\raisebox{-0.15em}{\includegraphics[height=1em]{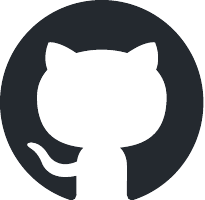}}}

\RequirePackage[
  datamodel=acmdatamodel,
  style=acmnumeric,
  ]{biblatex}

\usepackage{algorithm}
\usepackage{algpseudocode}
\usepackage{amsmath}
\usepackage{amsthm}

\usepackage[version=4]{mhchem}

\usepackage{multicol}
\usepackage{multirow}
\usepackage{array}
\usepackage{listings}

\begin{document}

\title{ARIA: A Causal-Aware Framework for Rescuing LLM Reasoning in Trustworthy Materials Discovery}
\author{Yi Cao}
\authornote{These authors contributed equally to this work.}
\affiliation{
  \department{Chemical and Biomolecular Engineering}
  \institution{Johns Hopkins University}
  \city{Baltimore}
  \state{MD}
  \country{USA}
}
\email{ycao73@jh.edu}

\author{Liaoyaqi Wang}
\authornotemark[1]
\affiliation{
  \department{Computer Science}
  \institution{Johns Hopkins University}
  \city{Baltimore}
  \state{MD}
  \country{USA}
}
\email{lwang240@jh.edu}

\author{Jieneng Chen}
\affiliation{
  \department{Computer Science}
  \institution{Johns Hopkins University}
  \city{Baltimore}
  \state{MD}
  \country{USA}
}
\email{jchen293@jhu.edu}

\author{Benjamin Van Durme}
\authornote{Corresponding author.}
\affiliation{
  \department{Computer Science}
  \institution{Johns Hopkins University}
  \city{Baltimore}
  \state{MD}
  \country{USA}
}
\email{vandurme@jhu.edu}

\author{Alan Yuille}
\authornotemark[2]
\affiliation{
  \department{Computer Science}
  \institution{Johns Hopkins University}
  \city{Baltimore}
  \state{MD}
  \country{USA}
}
\email{ayuille1@jhu.edu}

\author{Paulette Clancy}
\authornotemark[2]
\affiliation{
  \department{Chemical and Biomolecular Engineering}
  \institution{Johns Hopkins University}
  \city{Baltimore}
  \state{MD}
  \country{USA}
}
\email{pclancy3@jhu.edu}


\renewcommand{\shortauthors}{Yi Cao et al.}


\begin{abstract}
Generative models have revolutionized the process of materials discovery, yet they often fail to satisfy underlying physical causality.
Through an analysis of Large Language Models (LLMs) augmented with knowledge graphs derived from current literature, we uncover a phenomenon termed \textit{contextual tunneling}, where models ``over-anchor'' on narrow, retrieved evidence while suppressing global physical reasoning. 
To address this problem, we introduce \texttt{ARIA}, a causal-aware framework that conditions knowledge use on mechanistic completeness. \texttt{ARIA} routes each query through a three-tier cascade: (i) direct causal reasoning when complete evidence chains of Process-Structure-Property (PSP) are available, (ii) physics-informed analogical transfer for sparse or novel material systems, and (iii) explicit parametric fallback when external evidence is incomplete. As a proof of concept, we construct a Knowledge Graph (KG) containing 2,839 extracted PSP relations from peer-reviewed articles in the materials literature and evaluate \texttt{ARIA} on forward prediction and inverse design tasks for two-dimensional (2D) materials. \texttt{ARIA} mitigates contextual tunneling, improves over unaugmented and naive KG-augmented baselines, and provides further gains when an online literature search is used for evidence enrichment. Crucially, \texttt{ARIA} produces auditable causal traces, enabling physically grounded and trustworthy AI-assisted materials discovery.
\end{abstract}



\begin{teaserfigure}
\centering
    \textbf{\github ~ \textbf{Source Code:}  \url{https://github.com/yicao-elina/ARIA}} 
    \vspace{1em} 
    \includegraphics[width=\textwidth]{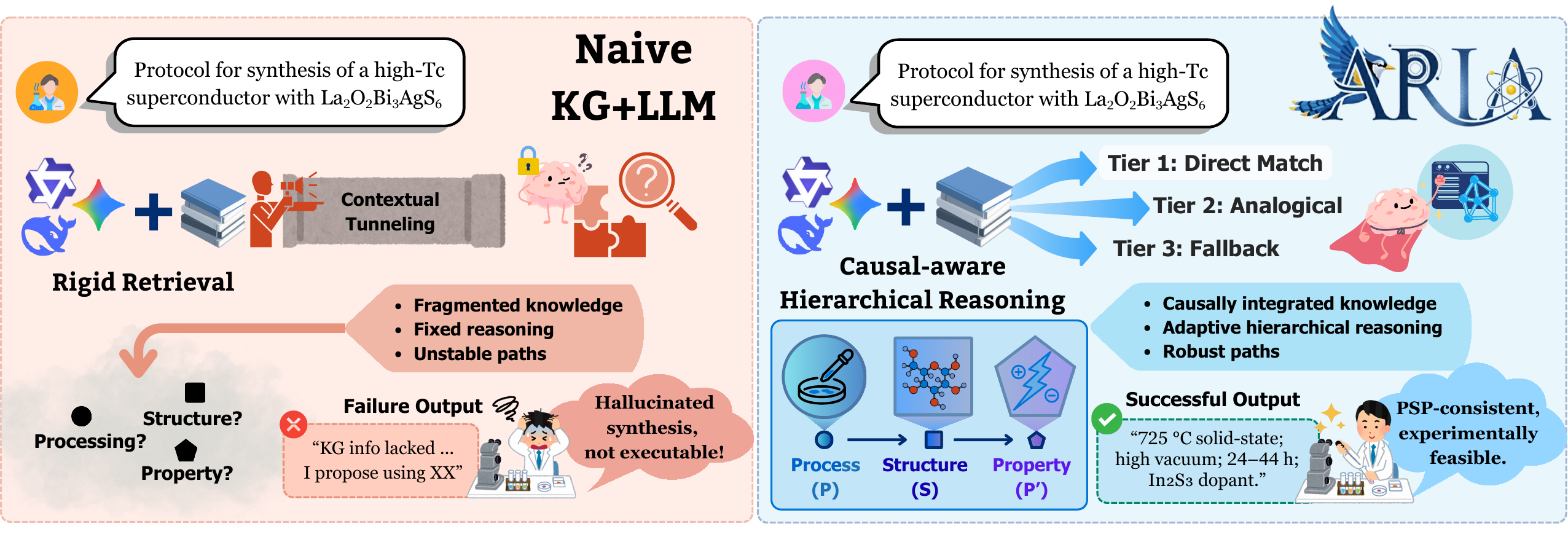}
    \caption{
    Naive Knowledge Graph-augmented LLMs can suffer from contextual tunneling (left); \texttt{ARIA} routes the same query through a three-tier adaptive cascade, mitigating this via hierarchical, physically constrained reasoning (right).
    }
    \label{fig:contextual_tunneling_vs_aria}
\end{teaserfigure}

\maketitle

\section{Introduction}
\label{sec:intro}

The pursuit of autonomous materials discovery, characterized by a closed feedback loop between generative design and experimental validation, is a central goal of computational materials science.~\cite{szymanski2023autonomous, ghareeb2026multi, lu2026towards, huang2026automat} In this paradigm, artificial intelligence (AI) helps navigate large chemical and processing spaces to propose synthesis protocols that are actionable in the laboratory. However, a fundamental gap remains: While AI systems can identify promising candidate materials, they often struggle to specify physically viable pathways for realizing them. This gap between \textit{in silico} design and wet-lab feasibility arises from the difficulty of reconciling broad linguistic knowledge with the causal constraints imposed by thermodynamics, chemical kinetics, phase stability, and synthesis-process windows.

Computational materials science has evolved from traditional machine learning models, which are often restricted to narrow structure--property correlations,~\cite{jain2013commentary,schmidt2019recent} toward Large Language Models (LLMs) capable of contextual reasoning and cross-domain synthesis.~\cite{jablonka2024leveraging, roy2026knowledge} Although LLMs can capture useful chemical patterns from the literature, they remain prone to mechanistic hallucinations: They may generate synthesis pathways that appear plausible from a text or linguistic point of view but violate chemical valency constraints, phase stability, or physically feasible transformation routes.~\cite{zhang2025large} For experimentalists, this creates a trust barrier in practice. Without a mechanism for auditing whether a model's reasoning follows established physical principles, AI-generated synthesis plans risk being persuasive but experimentally unusable.

Bridging this divide requires representing literature evidence in a form that exposes physical causal structure rather than relying solely on unstructured text. We argue that the Processing--Structure--Property (PSP) paradigm~\cite{olson1997computational,butler_machine_2018,schmidt2019recent} provides a natural architecture for this purpose. Instead of injecting raw literature passages into a model, the PSP framework distills synthesis knowledge into causal chains linking processing conditions to structural outcomes and functional properties. Such structured evidence can provide LLMs with recent experimental findings while emphasizing the mechanisms by which processing parameters influence material performance.

However, integrating structured Knowledge Graphs (KGs)~\cite{wang2024biorag,bazgir2025proteinhypothesis} with LLMs introduces a non-trivial challenge. In naive retrieval-augmented generation (RAG), retrieved KG evidence is typically provided to the LLM based on semantic or graph proximity, without assessing whether the evidence completes the causal chain required by the task. This can paradoxically degrade reasoning: the model may over-anchor on a locally relevant but causally incomplete fragment, such as a single Processing$\rightarrow$Structure relation, while neglecting the broader PSP pathway needed for synthesis prediction or design. We term this failure mode as \textbf{contextual tunneling}.

This pattern appears in our benchmarks. On in-domain forward prediction, Naive KG+LLM fails to improve over the unaugmented Baseline LLM, achieving an overall score of $0.337{\pm}0.027$ compared with $0.340{\pm}0.033$ for the Baseline LLM. Although the difference is small, it illustrates a broader failure mode. Providing retrieved evidence without signaling whether it covers the full causal pathway can constrain reasoning rather than enhance it. For example, evidence that ``high temperature improves crystallinity'' may be locally correct, but it is insufficient for synthesis design unless connected to precursor chemistry, phase formation, defect evolution, and the target property. In this sense, the problem is not lack of knowledge access, but lack of evidence sufficiency control.

To address contextual tunneling failure modes, we introduce \texttt{ARIA} (\textbf{A}utonomous \textbf{R}easoning \textbf{I}ntelligence for \textbf{A}ccelerated material discovery), a framework that regulates knowledge use by requiring retrieved evidence to satisfy a causal completeness criterion. Specifically, evidence should support a meaningful path in the Processing--Structure--Property hierarchy before being activated for model reasoning. \texttt{ARIA} implements a three-tier adaptive cascade:

\begin{enumerate}
    \item \textbf{Tier 1 --- Direct Causal Reasoning}: When a complete PSP path exists in the knowledge graph, such as \\
    Processing$\rightarrow$Structure$\rightarrow$Property, \texttt{ARIA} retrieves and uses this evidence directly. This restricts the LLM to evidence that provides a complete mechanistic picture rather than isolated fragments.
    \item \textbf{Tier 2 --- Physics-Informed Analogical Transfer}: When no complete PSP path exists for the queried material, \texttt{ARIA} identifies structurally or chemically similar materials in the KG and transfers their mechanistic PSP pathways, subject to physical feasibility constraints such as composition compatibility and thermal stability windows.
    \item \textbf{Tier 3 --- Parametric Fallback}: When neither direct evidence nor valid analogies are available, \texttt{ARIA} explicitly signals that external evidence is incomplete and allows the LLM to rely on its parametric physical knowledge while flagging reduced confidence.
\end{enumerate}

The key insight is that \textit{selective} evidence activation, gated by causal completeness, allows knowledge to enhance rather than constrain reasoning. By using retrieved evidence only when it provides a sufficiently complete mechanistic pathway, \texttt{ARIA} preserves the LLM's capacity for global physical reasoning while leveraging domain-specific evidence when it is causally informative.

Evaluated on expert-validated materials synthesis tasks spanning forward prediction and inverse design, \texttt{ARIA} delivers three main contributions: (1) \textbf{Performance gains}: improving over unaugmented LLM and naive KG-augmented baselines in both forward prediction and inverse design; (2) \textbf{Mechanistic accountability}: producing auditable causal traces that allow scientists to inspect the physical plausibility of model reasoning and (3) \textbf{Robust generalization}: maintaining performance on out-of-domain materials through physics-informed analogical transfer. Together, these establish a physically grounded framework for trustworthy, knowledge-guided materials discovery.

\begin{figure}
    \centering
    \small
    \includegraphics[width=\linewidth]{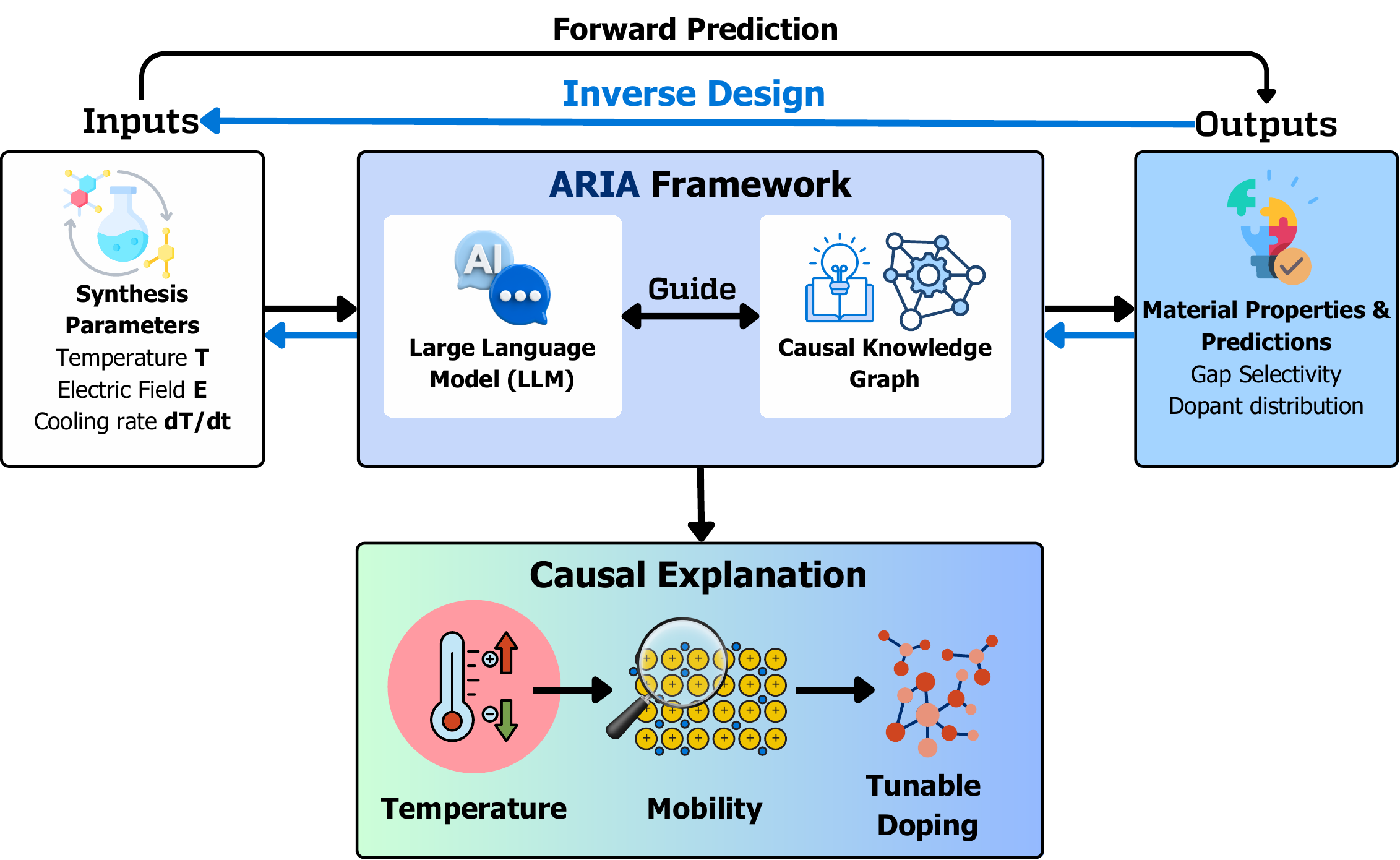}
    \caption{Schematic of the \texttt{ARIA} framework for bidirectional reasoning in materials discovery. The framework predicts material properties from synthesis parameters in forward tasks, while enabling inverse design by generating synthesis protocols from target properties. 
    }
    \label{fig:materials discovery}
\end{figure}
\section{Related Work}
\label{sec:related_work}

\paragraph{Computational Materials Discovery and the PSP Paradigm. }
Materials informatics has evolved from high-throughput Density Functional Theory (DFT) calculations~\cite{jain2013commentary, curtarolo2013high} to second-generation inverse design using differentiable surrogate models.~\cite{kim_inorganic_2020, noh2019inverse} While these methods enable rapid screening, they often operate on isolated segments of the \textbf{Processing--Structure--Property (PSP)} hierarchy.~\cite{olson1997computational, butler_machine_2018,raguraman2025call} Recent ``third-generation'' approaches utilize autonomous experimentation and Bayesian optimization,~\cite{szymanski2023autonomous, burger2020mobile, priyadarshini2024pal} yet they remain constrained by high-dimensional search spaces and the lack of a unified reasoning framework that captures the full causal chain from synthesis to electronic behavior.~\cite{li2022using,schmidt2019recent}

\paragraph{LLMs in Scientific Inquiry.}
 
Large Language Models (LLMs)~\cite{brown2020language} have emerged as promising engines for scientific synthesis.~\cite{xu2025largereasoningmodelssurvey} In materials science, domain-specific models such as MatSciBERT~\cite{gupta_matscibert_2022} and MatBERT~\cite{trewartha2022quantifying} improve entity recognition but struggle with the "mechanistic propagation" of physical laws.~\cite{jiang_applications_2025} While prompt-based systems like ChemCrow~\cite{bran_chemcrow_2023} and MatChat~\cite{chen_matchat_2023} demonstrate zero-shot utility, they frequently generate synthesis protocols that violate fundamental thermodynamic constraints or exhibit poor numerical precision in property prediction.~\cite{dwhite_assessment_2023, dagdelen_structured_2024} The core limitation remains that unstructured text lacks explicit encoding of the causal directionality required for rigorous scientific discovery.~\cite{kiciman2023causal, zheng_chatgpt_2023}

\paragraph{Knowledge Graphs and the Paradox of Augmentation.}
 Knowledge Graphs (KGs) provide the structured, directional evidence required for causal inference~\cite{zhang2024causalgraphdiscoveryretrievalaugmented, zheng2025large}. In chemistry and biology, KG-LLM integration has improved factual grounding through retrieval-augmented generation (RAG)~\cite{10.5555/3495724.3496517, wang2024biorag}. However, recent studies identify a "Knowledge Paradox" in which external augmentation does not invariably improve reasoning. Several studies \cite{longpre-etal-2021-entity, xu-etal-2024-knowledge-conflicts} observe that LLMs often over-anchor on retrieved context even when it contradicts their internal parametric prior. While frameworks like GIVE~\cite{he2024give} and GNN-RAG~\cite{edge2025localglobalgraphrag} attempt to mitigate noise in general-purpose KGs, they do not address the specific challenge of \textit{mechanistic incompleteness} in scientific domains, the case where retrieved evidence is correct but causally narrow.

\paragraph{Contextual Tunneling as a Distinct Failure Mode.}

This paper identifies and empirically demonstrates \textbf{contextual tunneling}, a failure mode wherein naive KG-augmented LLMs over-anchor on narrow, correct evidence while suppressing broader physical reasoning. Unlike the "Knowledge Paradox," which describes contradiction between retrieved and parametric knowledge, contextual tunneling occurs even when retrieved evidence is factually accurate. We document this empirically using materials science tasks: naive KG+LLM integration achieves lower scores than unaugmented LLM in forward prediction (Naive KG+LLM: 0.337$\pm$0.027 vs. Baseline LLM: 0.340$\pm$0.033), demonstrating that undiscriminating knowledge integration can degrade rather than enhance reasoning.

To address this gap, we introduce \texttt{ARIA}, which employs \textit{inference-time adaptive gating} based on evidence completeness: selectively activating structured KG reasoning only when the retrieved evidence spans a complete causal path in the Processing-Structure-Property hierarchy. By conditioning knowledge use on mechanistic completeness rather than retrieval confidence alone, \texttt{ARIA} ensures that knowledge enhances—rather than constrains—scientific discovery.~\cite{amayuelas2025groundingllmreasoningknowledge, yoran2024making} This approach is distinct from existing noise-mitigation and ranking-based augmentation strategies, which operate primarily at the retrieval stage whereas \texttt{ARIA} gates knowledge activation at the reasoning stage based on mechanistic criteria.
\section{Method}
\label{sec:method}

\begin{figure*}[ht]
\centering
\includegraphics[width=0.7\linewidth]{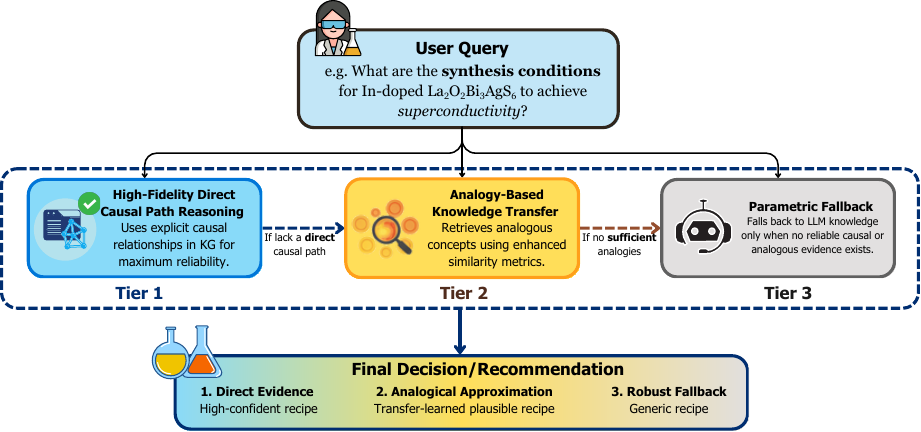} 
\caption{\textbf{Schematic of the} \texttt{ARIA} \textbf{Model Architecture}}
\label{fig:ARIA}
\end{figure*}

The method consists of three components. We first formalize forward prediction and inverse design as PSP causal reasoning tasks (\autoref{sec:psp_tasks}). We then define the PSP-completeness criterion used for tier selection (\autoref{sec:contextual_tunneling}) and describe the Causal Knowledge Graph that stores literature-derived PSP relations (\autoref{sec:causal_kg}). Finally, we present the three-tier inference cascade that combines direct causal reasoning, constraint-aware analogical transfer, and parametric fallback (\autoref{sec:aria_arch}).

\subsection{Materials Discovery as PSP Causal Reasoning}
\label{sec:psp_tasks}

Materials discovery requires reasoning over the causal chain
\textit{processing} $\rightarrow$ \textit{structure} $\rightarrow$ \textit{property} (PSP)~\cite{olson1997computational}.
We study two tasks: (i) \textbf{forward prediction} (synthesis conditions $\mathcal{S}$ $\rightarrow$ properties $\mathcal{P}$) and (ii) \textbf{inverse design} (target properties $\mathcal{P}^*$ $\rightarrow$ synthesis conditions $\mathcal{S}^*$), both under feasibility constraints (\textit{e.g.,} stability windows, precursor compatibility).
The full formalization (objective functions, constraint sets, and examples) is provided in the Supplementary Information (SI).

\subsection{Contextual Tunneling in PSP Reasoning}
\label{sec:contextual_tunneling}

To make contextual tunneling operational, we define when retrieved evidence is PSP-complete.
For example, suppose the KG contains evidence that ``high temperature improves crystallinity'' (a processing--structure link). Naive RAG passes this to the LLM without indicating that the structure--property link is missing. The LLM then over-anchors on this narrow mechanism and loses sight of broader thermodynamic constraints. This degradation is \textit{not} due to false information, but to mechanistic incompleteness.

\paragraph{Formal criterion.}
\label{def:psp_complete}
To operationalize completeness, let $\mathcal{G} = (\mathcal{V}, \mathcal{E})$ be the knowledge graph with vertex partition
$\mathcal{V} = \mathcal{V}_P \cup \mathcal{V}_S \cup \mathcal{V}_{\mathrm{Prop}}$
(synthesis procedures, host systems, and properties), and edge confidence weights $w(e)\in[0,1]$.
For a query $q = (\textit{method}, \textit{host}, \textit{dopant})$ with induced subgraph $\mathcal{G}_q \subseteq \mathcal{G}$,
we say $q$ is \emph{PSP-complete} if there exists a three-hop path
$p = (v_P,\, v_S,\, v_{\mathrm{Prop}})$ in $\mathcal{G}_q$ satisfying:
(i)~$v_P\!\in\!\mathcal{V}_P$ matches the synthesis method;
(ii)~$v_S\!\in\!\mathcal{V}_S$ matches the host/dopant;
(iii)~$(v_P,v_S),(v_S,v_{\mathrm{Prop}})\in\mathcal{E}$; and
(iv)~$\min_{e\in p}w(e)\ge\theta_1$.

\paragraph{Tier activation.}
PSP-completeness directly governs \texttt{ARIA}'s reasoning tier:
\begin{equation}
\text{Tier} =
\begin{cases}
1 & \text{if } \mathrm{PSP\text{-}complete}(q) = \mathrm{True} \\[2pt]
2 & \text{if } \neg\,\mathrm{PSP\text{-}complete}(q)\ \wedge\ \max_{q'}\mathrm{sim}(q,q') \ge \theta_2 \\[2pt]
3 & \text{otherwise (parametric fallback)}
\end{cases}
\label{eq:tier_selection}
\end{equation}

\noindent
\texttt{ARIA} prevents contextual tunneling through this inference-time adaptive gating: structured retrieval activates only when evidence satisfies PSP-completeness; otherwise, the LLM reasons from parametric knowledge without the risk of anchoring on a causally incomplete fragment. 

\subsection{Causal Knowledge Graph for PSP Relationships}
\label{sec:causal_kg}

To ground \texttt{ARIA}'s reasoning in verifiable PSP mechanisms, we construct a Causal Knowledge Graph $\mathcal{G} = (\mathcal{V}, \mathcal{E})$ encoding literature-verified relationships between processing conditions, structural features, and material properties.

\paragraph{Construction Pipeline.}
We extract PSP entities and causal relations from a materials corpus and assemble a directed graph $\mathcal{G}=(\mathcal{V},\mathcal{E})$ whose edges encode literature-supported, directionally consistent PSP links.
To reduce spurious edges, we apply constraint-aware filtering (\textit{e.g.,} basic thermodynamic and stoichiometric checks) and retain provenance for every relation.
Full extraction prompts, ontologies, and filtering criteria are provided in the SI.

The resulting graph (\autoref{fig:KG}) enables traversal from processing conditions to structural mechanisms and target properties.

\subsection{\texttt{ARIA} Architecture: Three-Tier Adaptive Reasoning}
\label{sec:aria_arch}

\texttt{ARIA} prevents contextual tunneling through a three-tier cascade (\autoref{fig:ARIA}) that uses structured evidence only when it is sufficiently complete.
Algorithmic pseudocode for the full cascade and its tier-specific procedures is provided in ~\autoref{app:algorithm}, including Algorithm \ref{alg:aria} (overall flow), Algorithm \ref{alg:tier1} (Tier 1), and Algorithm \ref{alg:tier2} (Tier 2).

\paragraph{Tier 1: Direct causal reasoning.}
If a complete PSP path exists in $\mathcal{G}$ for the query entities (i.e., the knowledge graph contains verified links spanning processing $\rightarrow$ structure $\rightarrow$ property), \texttt{ARIA} traverses the KG, aggregates all consistent paths, and prompts the LLM to generate an answer grounded in those mechanistically complete mechanisms. 
\textit{Why this prevents fixation}: By restricting evidence to complete PSP chains, Tier 1 ensures the LLM receives a full causal narrative rather than a narrow processing-structure fragment, preventing over-anchoring on incomplete mechanisms.

\paragraph{Tier 2: Physics-informed analogy.}
If no direct complete path exists (\textit{e.g.,} for a novel material composition), \texttt{ARIA} retrieves the most structurally and compositionally similar materials in $\mathcal{G}$ and transfers their PSP mechanistic pathways, subject to lightweight physical feasibility checks (\textit{e.g.,} structural class compatibility, elemental substitution rules, thermal stability windows). 
\textit{Why this preserves validity}: Physical constraints gate the transfer, ensuring analogical evidence only activates when mechanistically justified, preventing phantom analogies that violate physical laws.

\paragraph{Tier 3: Parametric fallback.}
If neither direct evidence nor credible analogies exist, \texttt{ARIA} \textit{disables retrieval} and relies on the LLM's parametric knowledge, explicitly flagging to the user that the output is speculative and not grounded in domain evidence.
\textit{Why this prevents hallucination}: By honestly signaling evidence absence rather than providing weak or incomplete evidence, Tier 3 prevents the LLM from implicitly anchoring to false evidence. Users can then verify the output against known physical laws without being misled by apparent grounding.

Algorithmic details (tier selection, scoring, and prompts) and hyperparameters are provided in the \autoref{app:algorithm};
The representative case studies for each tier are included in the tutorial in 
\hyperlink{https://github.com/yicao-elina/ARIA.git}{ARIA GitHub repository}.

\paragraph{Why Adaptive Tiering Prevents Contextual Tunneling.}

The three-tier cascade implements evidence completeness assessment at the algorithmic level. In naive RAG, the LLM receives all retrieved KG subgraphs without a mechanism to determine whether they form complete causal paths. This can lead to over-anchoring: the LLM conditions heavily on narrow evidence (\textit{e.g.,} a processing-structure link) while inadvertently suppressing its knowledge of other causal links (\textit{e.g.,} structure-property constraints). 

\texttt{ARIA} prevents this by refusing to provide evidence unless it satisfies a completeness criterion: Tier 1 enforces that evidence spans the full processing $\rightarrow$ structure $\rightarrow$ property chain, Tier 2 gates analogical transfer by physical constraints, and Tier 3 explicitly signals when no evidence meets these criteria, permitting the LLM to rely on its parametric knowledge without false anchoring. The key insight is that \textit{selective} evidence activation, gated by causal completeness rather than retrieval confidence alone, allows knowledge to enhance rather than constrain reasoning.


\section{Experiments}
\label{sec:experiments}

\begin{figure*}[h]
    \centering
    \includegraphics[width=0.85\linewidth]{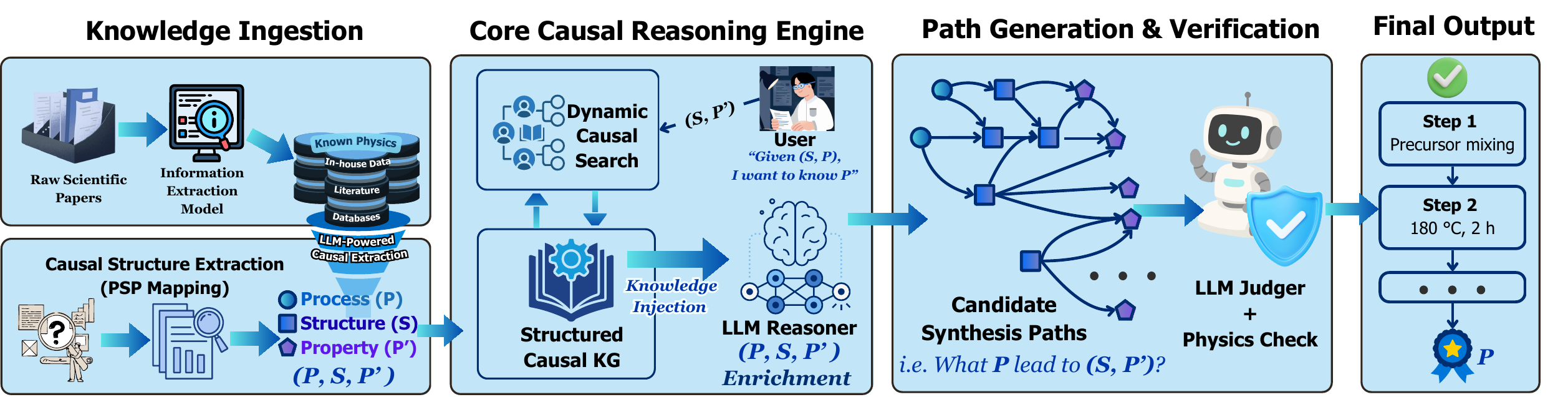}
\caption{\textbf{Knowledge graph construction pipeline and workflow for materials processing pathway prediction.}}
    \label{fig:KG}
\end{figure*}

We evaluate the central hypothesis of \texttt{ARIA}: knowledge-augmented LLM reasoning in materials science improves when retrieved evidence is activated only after satisfying a causal completeness criterion, defined by coverage of the Processing--Structure--Property (PSP) chain.
Our experiments address two primary research questions:

\begin{enumerate}
    \item[(Q1)] Does naive KG integration (without adaptive gating) systematically underperform compared to unaugmented LLM reasoning?
    \item[(Q2)] Does \texttt{ARIA}'s tier-based evidence filtering recover performance lost to naive integration and further improve it?
\end{enumerate}

\subsection{Experimental Setup}

\paragraph{Knowledge Graph Construction.}

We constructed a Causal Knowledge Graph from peer-reviewed materials science literature published between 2005 and 2023 (\autoref{fig:KG}). The graph focuses on two-dimensional (2D) materials synthesis, structural evolution, and functional characterization. Our data collection pipeline comprised:

\begin{enumerate}
    \item \textbf{Literature Corpus}: We collected and processed 1,531 full-text PDF papers from high-impact materials science journals, together with 10,000 abstracts from a broader literature search on 2D materials synthesis and characterization.
    
    \item \textbf{Text Extraction and Preprocessing}: We extracted abstracts, experimental sections, synthesis descriptions, and characterization passages using automated text-mining pipelines, followed by domain-specific cleaning and normalization.
    
    \item \textbf{PSP Relation Extraction}: We applied Qwen2.5:7B to extract Processing--Structure--Property relationships from the processed paper content, yielding approximately 2,839 PSP relations, where training/graph-construction split into approximately 2,271 relations with a held-out relation split of approximately 568 relations.
    
    \item \textbf{Graph Assembly}: We constructed a directed multigraph encoding Processing$\rightarrow$Structure (P$\rightarrow$S) and Structure$\rightarrow$Property (S$\rightarrow$P) relationships. Edges were assigned confidence weights based on extraction consistency, source metadata, and agreement across related evidence statements.
\end{enumerate}

The resulting Causal Knowledge Graph spans 2D materials synthesis across diverse processing methods (CVD, sol-gel, hydrothermal, mechanical exfoliation, \textit{etc.}), structural descriptors (phase, crystal symmetry, thickness, defect density) and properties (electrical, thermal, optical, mechanical).

\paragraph{Test Set and Ground Truth.}

We curated a materials science evaluation set from peer-reviewed literature (2005-2026) focusing on 2D materials. Test cases were extracted and validated by materials scientists with expertise in crystal chemistry and functional property prediction. The final dataset comprises 149 expert-validated materials synthesis and design tasks:

\begin{itemize}
    \item \textbf{In-Domain} (117 cases): Materials with established literature precedent in the KG, split into:
    \begin{itemize}
        \item Forward prediction (58 cases): Given synthesis parameters (temperature, precursors, atmosphere), predict resulting structure and properties
        \item Inverse design (59 cases): Given target properties (\textit{e.g.,} band gap, conductivity), generate viable synthesis protocols
    \end{itemize}
    \item \textbf{Out-of-Domain} (32 cases): Novel material compositions and synthesis settings sourced from 2024--2026 publications that post-date the KG construction corpus.
\end{itemize}

Evaluation ground truth was sourced from: (1) experimental synthesis procedures and measured properties reported in source papers, (2) computational databases (Materials Project, AFLOW), and (3) standardized characterization protocols from materials science literature. Inter-annotator disagreement on task relevance was resolved by a consensus discussion.

\paragraph{System Variants.}

We implemented and evaluated six system configurations using DeepSeek-R1:8B as the base LLM for material prediction and inverse design tasks. Retrieval-based variants used the same embedding model, all-MiniLM-L6-v2, unless otherwise specified.

\begin{enumerate}
    \item \textbf{Baseline LLM}: DeepSeek-R1:8B without external knowledge sources. 
    
    \item \textbf{KG-Only}: A non-generative retrieval baseline that returns the most relevant KG relations or paths without LLM-based synthesis of the final answer. 
    
    \item \textbf{Naive KG+LLM}: DeepSeek-R1:8B with direct KG retrieval. Relevant KG subgraphs, including entities and edges matching the query, are retrieved and concatenated into the LLM prompt without causal completeness filtering or evidence sufficiency assessment. This baseline tests whether unfiltered knowledge integration helps or harms reasoning.
    
    \item \textbf{\texttt{ARIA}-CORE}: The three-tier \texttt{ARIA} cascade using only the static KG. No online search or external enrichment is used. 
    
    \item \textbf{\texttt{ARIA}-SEARCH}: A search-augmented variant that uses real-time literature search without the full tier-based KG reasoning cascade. 
    
    \item \textbf{\texttt{ARIA}-FULL}: The full framework, combining tier-based causal KG reasoning with real-time literature search augmentation. Search results are used to enrich the evidence base when static KG evidence is incomplete or absent.
\end{enumerate}

\paragraph{Evaluation Metrics and Protocol.}

We assess all outputs using automated constraint validation and structured LLM-as-judge scoring:

\begin{itemize}
    \item \textbf{Scientific Accuracy} (0-1): Adherence to physical principles and consistency with literature consensus
    
    \item \textbf{Functional Equivalence} (0-1): Whether proposed synthesis parameters achieve specified target properties
    
    \item \textbf{Completeness} (0-1): Specification of all required synthesis steps, parameter ranges, and auxiliary details (\textit{e.g.,} atmosphere control)
    
    \item \textbf{Interpretability} (0-1): Clarity of mechanistic reasoning and traceability of causal claims to specific synthesis steps
    
    \item \textbf{Overall Score} (0-1): Composite harmonic aggregate across the four dimensions to emphasize scientific validity
\end{itemize}

Automated checks include thermodynamic feasibility, stoichiometric consistency, and mass conservation. Thermodynamic feasibility is evaluated by comparing proposed synthesis temperatures with material stability constraints obtained from the Materials Project API when available. Textual quality and rubric-based scores are evaluated using DeepSeek-R1:14B, which is prompted with the same structured evaluation rubric across all systems.

\paragraph{Implementation Details.}

\textbf{Prediction and Design Model}: \\ DeepSeek-R1:8B serves as the base LLM for all forward prediction and inverse design tasks. The Causal Knowledge Graph was constructed using Qwen2.5:7B for automated extraction of PSP relationships from scientific papers.

\textbf{Evaluation Model}: DeepSeek-R1:14B evaluates output quality across the metrics described above using fixed prompts and deterministic decoding. We report dimension-specific scores together with a separate holistic overall score, which is assigned by the evaluator as an independent judgment.

\textbf{\texttt{ARIA} Hyperparameters}: Tier 1 activation requires evidence covering all three PSP dimensions relevant to the query. For Tier 2 analogical reasoning, we use an analogy similarity threshold of $\tau=0.75$ and retrieve the top $K=5$ candidate analogies. The analogy score is computed as a weighted combination of semantic, categorical, and numerical similarity, with weights $w_1=0.5$, $w_2=0.3$, and $w_3=0.2$, respectively.

\textbf{Retrieval}: For KG retrieval, we retrieve the top $k=3$ neighboring entities or relations per query using cosine similarity over all-MiniLM-L6-v2 embeddings, with a retrieval threshold of $\theta=0.65$. For Tier 2 analogical reasoning, structural similarity is computed from materials descriptors, including composition, crystal class, lattice type, and available processing constraints.

\textbf{Reproducibility}: Complete prompts, evaluation rubrics, hyperparameters, and implementation details are provided in the Supplementary Information.

\begin{table*}[ht!]
\centering
\caption{\textbf{In-domain vs. out-of-domain performance analysis.} 
We evaluate six systems on in-domain data (materials/protocols exist in KG) and out-of-domain data (novel materials/protocols not in KG). ARIA demonstrates improved overall performance across both forward prediction and inverse design tasks using physics-guided metrics.}
\label{tab:domain_generalization_analysis_v2}
\resizebox{0.92\textwidth}{!}{ 
\begin{tabular}{lcccccc}
\toprule
\textbf{System} & \textbf{Domain} & \textbf{Sci. Accuracy} & \textbf{Funct. Equiv.} & \textbf{Completeness} & \textbf{Interpretability} & \textbf{Overall Score} $\uparrow$ \\
\midrule
\multicolumn{7}{c}{\textbf{Forward Prediction}} \\
\midrule
Baseline LLM & In-Domain & 0.252±0.151 & 0.385±0.124 & 0.659±0.076 & 0.877±0.046 & \textbf{0.340±0.033} \\
Baseline LLM & Out-of-Domain & 0.237±0.181 & 0.344±0.149 & 0.649±0.067 & 0.883±0.031 & \textbf{0.326±0.034} \\
\textit{Domain Gap} & & \textcolor{red}{\textit{-6.0\%}} & \textcolor{red}{\textit{-10.6\%}} & \textcolor{red}{\textit{-1.5\%}} & \textcolor{blue}{\textit{+0.7\%}} & \textcolor{red}{\textit{-4.1\%}} \\
\cmidrule{1-7}
NAIVE-KG & In-Domain & 0.280±0.158 & 0.394±0.128 & 0.732±0.044 & 0.350±0.000 & \textbf{0.337±0.027} \\
NAIVE-KG & Out-of-Domain & 0.226±0.165 & 0.341±0.140 & 0.735±0.041 & 0.350±0.000 & \textbf{0.322±0.035} \\
\textit{Domain Gap} & & \textcolor{red}{\textit{-19.3\%}} & \textcolor{red}{\textit{-13.5\%}} & \textcolor{blue}{\textit{+0.4\%}} & \textit{0.0\%} & \textcolor{red}{\textit{-4.5\%}} \\
\cmidrule{1-7}
ARIA-CORE & In-Domain & 0.235±0.129 & 0.400±0.100 & 0.749±0.011 & 0.791±0.034 & \textbf{0.410±0.024} \\
ARIA-CORE & Out-of-Domain & 0.212±0.158 & 0.348±0.120 & 0.748±0.010 & 0.799±0.035 & \textbf{0.447±0.033} \\
\textit{Domain Gap} & & \textcolor{red}{\textit{-9.8\%}} & \textcolor{red}{\textit{-13.0\%}} & \textcolor{red}{\textit{-0.1\%}} & \textcolor{blue}{\textit{+1.0\%}} & \textcolor{red}{\textit{-1.5\%}} \\
\cmidrule{1-7}
ARIA-FULL & In-Domain & 0.254±0.152 & 0.339±0.092 & 0.930±0.026 & 0.925±0.000 & \textbf{0.512±0.039} \\
ARIA-FULL & Out-of-Domain & 0.205±0.167 & 0.278±0.111 & 0.918±0.032 & 0.925±0.000 & \textbf{0.513±0.031} \\
\textit{Domain Gap} & & \textcolor{red}{\textit{-19.3\%}} & \textcolor{red}{\textit{-18.0\%}} & \textcolor{red}{\textit{-1.3\%}} & \textit{0.0\%} & \textcolor{red}{\textit{+9.0\%}} \\
\cmidrule{1-7}
ARIA-SEARCH & In-Domain & 0.275±0.157 & 0.382±0.118 & 0.799±0.004 & 0.816±0.050 & \textbf{0.511±0.042} \\
ARIA-SEARCH & Out-of-Domain & 0.228±0.173 & 0.347±0.159 & 0.799±0.006 & 0.805±0.038 & \textbf{0.493±0.039} \\
\textit{Domain Gap} & & \textcolor{red}{\textit{-17.1\%}} & \textcolor{red}{\textit{-9.2\%}} & \textit{0.0\%} & \textcolor{red}{\textit{-1.3\%}} & \textcolor{red}{\textit{-3.5\%}} \\
\cmidrule{1-7}
KG-ONLY & In-Domain & 0.068±0.049 & 0.143±0.054 & 0.600±0.000 & 0.350±0.000 & \textbf{0.231±0.009} \\
\cmidrule{1-7}
\multicolumn{7}{c}{\textbf{Performance Comparison}} \\
\midrule
NAIVE-KG vs Baseline &  & \textcolor{red}{\textit{-4.6\%}} & \textcolor{red}{\textit{-0.9\%}} & \textcolor{blue}{\textit{+13.3\%}} & \textcolor{red}{\textit{-60.4\%}} & \textcolor{red}{\textit{-0.9\%}} \\
ARIA-CORE vs Baseline &  & \textcolor{red}{\textit{-10.5\%}} & \textcolor{blue}{\textit{+1.2\%}} & \textcolor{blue}{\textit{+15.3\%}} & \textcolor{red}{\textit{-9.5\%}} & \textcolor{blue}{\textit{+20.6\%}} \\
ARIA-FULL vs Baseline &  & \textcolor{red}{\textit{-13.5\%}} & \textcolor{red}{\textit{-19.2\%}} & \textcolor{blue}{\textit{+41.4\%}} & \textcolor{blue}{\textit{+4.8\%}} & \textcolor{blue}{\textit{+50.6\%}} \\
ARIA-SEARCH vs Baseline &  & \textcolor{red}{\textit{-3.8\%}} & \textcolor{blue}{\textit{+0.9\%}} & \textcolor{blue}{\textit{+23.1\%}} & \textcolor{red}{\textit{-8.8\%}} & \textcolor{blue}{\textit{+50.3\%}} \\
KG-ONLY vs Baseline &  & \textcolor{red}{\textit{-75.5\%}} & \textcolor{red}{\textit{-64.5\%}} & \textcolor{red}{\textit{-7.6\%}} & \textcolor{red}{\textit{-60.4\%}} & \textcolor{red}{\textit{-32.1\%}} \\
\midrule
\multicolumn{7}{c}{\textbf{Inverse Design}} \\
\midrule
\textbf{System} & \textbf{Domain} & \textbf{Sci. Accuracy} & \textbf{Funct. Equiv.} & \textbf{Completeness} & \textbf{Interpretability} & \textbf{Overall Score} \\
\midrule
Baseline LLM & In-Domain & 0.265±0.147 & 0.507±0.085 & 0.746±0.012 & 0.831±0.017 & \textbf{0.345±0.018} \\
Baseline LLM & Out-of-Domain & 0.303±0.197 & 0.448±0.163 & 0.707±0.072 & 0.587±0.361 & \textbf{0.313±0.047} \\
\textit{Domain Gap} & & \textcolor{red}{\textit{+14.3\%}} & \textcolor{red}{\textit{-11.6\%}} & \textcolor{red}{\textit{-5.2\%}} & \textcolor{red}{\textit{-29.3\%}} & \textcolor{red}{\textit{-9.3\%}} \\
\cmidrule{1-7}
NAIVE-KG & In-Domain & 0.186±0.068 & 0.465±0.169 & 0.724±0.078 & 0.350±0.000 & \textbf{0.301±0.047} \\
NAIVE-KG & Out-of-Domain & 0.277±0.191 & 0.478±0.166 & 0.745±0.068 & 0.350±0.000 & \textbf{0.312±0.046} \\
\textit{Domain Gap} & & \textcolor{blue}{\textit{+48.9\%}} & \textcolor{blue}{\textit{+2.8\%}} & \textcolor{blue}{\textit{+2.9\%}} & \textit{0.0\%} & \textcolor{blue}{\textit{+3.7\%}} \\
\cmidrule{1-7}
ARIA-CORE & In-Domain & 0.205±0.032 & 0.513±0.080 & 0.800±0.000 & 0.803±0.024 & \textbf{0.454±0.015} \\
ARIA-CORE & Out-of-Domain & 0.278±0.173 & 0.505±0.122 & 0.792±0.021 & 0.813±0.026 & \textbf{0.459±0.028} \\
\textit{Domain Gap} & & \textcolor{blue}{\textit{+35.6\%}} & \textcolor{red}{\textit{-1.6\%}} & \textcolor{red}{\textit{-1.0\%}} & \textcolor{blue}{\textit{+1.2\%}} & \textcolor{blue}{\textit{+1.1\%}} \\
\cmidrule{1-7}
ARIA-FULL & In-Domain & 0.250±0.156 & 0.395±0.073 & 0.803±0.020 & 0.925±0.000 & \textbf{0.498±0.034} \\
ARIA-FULL & Out-of-Domain & 0.398±0.205 & 0.424±0.071 & 0.805±0.020 & 0.925±0.000 & \textbf{0.513±0.040} \\
\textit{Domain Gap} & & \textcolor{blue}{\textit{+59.2\%}} & \textcolor{blue}{\textit{+7.3\%}} & \textcolor{blue}{\textit{+0.2\%}} & \textit{0.0\%} & \textcolor{blue}{\textit{+3.0\%}} \\
\cmidrule{1-7}
ARIA-SEARCH & In-Domain & 0.246±0.135 & 0.471±0.103 & 0.785±0.016 & 0.825±0.050 & \textbf{0.520±0.042} \\
ARIA-SEARCH & Out-of-Domain & 0.221±0.102 & 0.469±0.107 & 0.774±0.028 & 0.777±0.157 & \textbf{0.467±0.072} \\
\textit{Domain Gap} & & \textcolor{red}{\textit{-10.2\%}} & \textcolor{red}{\textit{-0.4\%}} & \textcolor{red}{\textit{-1.4\%}} & \textcolor{red}{\textit{-5.8\%}} & \textcolor{red}{\textit{-10.2\%}} \\
\cmidrule{1-7}
KG-ONLY & In-Domain & 0.080±0.023 & 0.201±0.056 & 0.600±0.000 & 0.350±0.000 & \textbf{0.238±0.008} \\
\cmidrule{1-7}
\multicolumn{7}{c}{\textbf{Performance Comparison}} \\
\midrule
NAIVE-KG vs Baseline &  & \textcolor{red}{\textit{-29.8\%}} & \textcolor{red}{\textit{-8.3\%}} & \textcolor{red}{\textit{-2.9\%}} & \textcolor{red}{\textit{-57.9\%}} & \textcolor{red}{\textit{\textbf{-12.7\%}}} \\
ARIA-CORE vs Baseline &  & \textcolor{red}{\textit{-22.6\%}} & \textcolor{blue}{\textit{+1.2\%}} & \textcolor{blue}{\textit{+7.2\%}} & \textcolor{red}{\textit{-3.4\%}} & \textcolor{blue}{\textit{\textbf{+31.6\%}}} \\
ARIA-FULL vs Baseline &  & \textcolor{red}{\textit{-5.7\%}} & \textcolor{red}{\textit{-22.1\%}} & \textcolor{blue}{\textit{+7.6\%}} & \textcolor{blue}{\textit{+11.3\%}} & \textcolor{blue}{\textit{\textbf{+44.3\%}}} \\
ARIA-SEARCH vs Baseline &  & \textcolor{red}{\textit{-7.2\%}} & \textcolor{red}{\textit{-7.1\%}} & \textcolor{blue}{\textit{+5.2\%}} & \textcolor{red}{\textit{-0.7\%}} & \textcolor{blue}{\textit{\textbf{+50.7\%}}} \\
KG-ONLY vs Baseline &  & \textcolor{red}{\textit{-69.8\%}} & \textcolor{red}{\textit{-60.4\%}} & \textcolor{red}{\textit{-19.6\%}} & \textcolor{red}{\textit{-57.9\%}} & \textcolor{red}{\textit{\textbf{-30.9\%}}} \\
\bottomrule
\end{tabular}}
\end{table*}

\begin{figure*}[ht]
\centering
\includegraphics[width=0.9\textwidth]{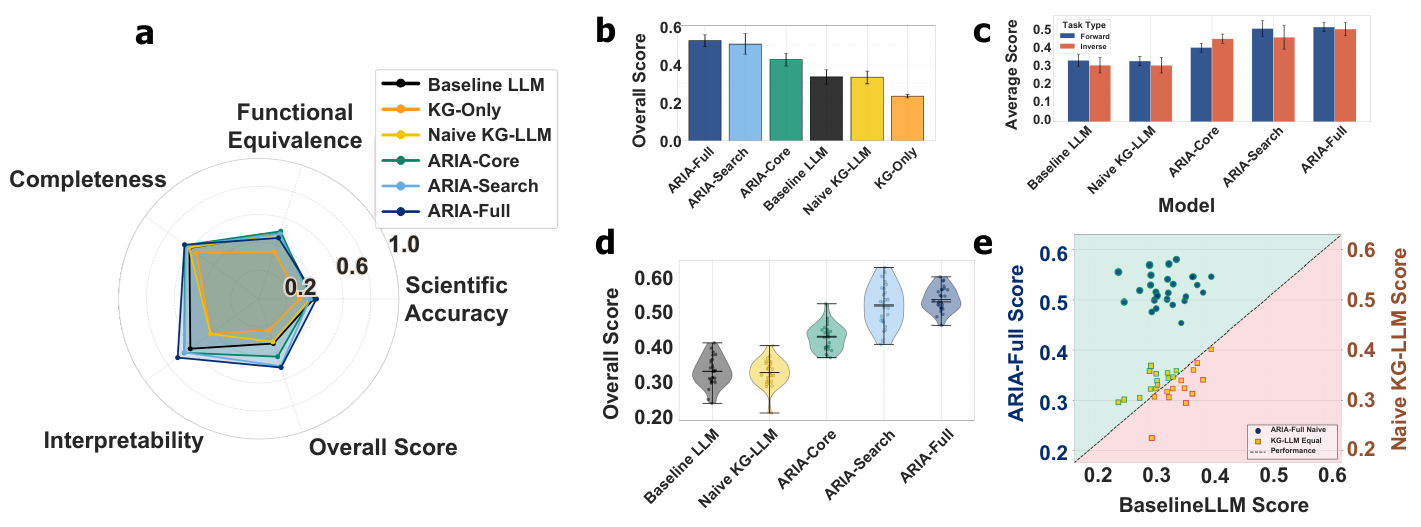}
\caption{
\textbf{Comprehensive evaluation of LLM-based scientific reasoning and design.}
(a) Radar plot summarizing performance across five evaluation metrics for all models (Baseline, KG-Only, Naive KG+LLM, \texttt{ARIA}-Core, \texttt{ARIA}-Search, \texttt{ARIA}-Full).
(b) Mean overall scores with standard deviation error bars.
(c) Violin plots showing overall score distributions.
(d) Performance by task type (Forward Prediction vs.\ Inverse Design).
(e) Scatter comparison of \texttt{ARIA}-Full (circles) and Naive KG+LLM (squares) against the Baseline, with green/red outlines indicating outperforming/underperforming cases ($y>x$ / $y<x$); marker size scales with deviation from the equality line.
}
\label{fig:test-evaluation}
\end{figure*}

\subsection{Main Results: In-Domain and Out-of-Domain Performance}

\noindent\textit{Finding 1: Naive KG integration does not improve over the unaugmented LLM.}

\noindent On in-domain forward prediction tasks, Naive KG+LLM achieves an overall score of $0.337{\pm}0.027$, compared to $0.340{\pm}0.033$ for the Baseline LLM. Although the difference is small, Naive KG retrieval fails to improve over purely parametric reasoning and slightly underperforms on average. This result is consistent with the contextual tunneling hypothesis: When retrieved evidence is injected without assessing causal completeness, the LLM can over-anchor on narrow evidence fragments, such as a single Processing$\rightarrow$Structure relation, rather than integrating evidence into a complete mechanistic PSP narrative. This motivates \texttt{ARIA}'s central design choice: Retrieval should be governed by evidence sufficiency rather than semantic relevance alone.\\

\noindent\textit{Finding 2: \texttt{ARIA}-CORE recovers performance through evidence gating.}

\noindent\texttt{ARIA}-CORE, which implements tier-based reasoning using the static KG without an online search, achieves an overall score of $0.410{\pm}0.024$ on in-domain forward prediction. This corresponds to a $+21.6\%$ improvement over Naive KG+LLM and a $+20.6\%$ improvement over the Baseline LLM. The recovery from naive retrieval indicates that causal completeness assessment is a key mechanism: Tier~1 restricts direct evidence use to complete PSP chains, while Tier~2 permits analogical transfer only when the analogy satisfies physical consistency constraints. Together, these gates reduce the risk that incomplete retrieved evidence narrows the model's reasoning.\\

\noindent\textit{Finding 3: \texttt{ARIA}-FULL high performance via literature integration.}

\noindent\texttt{ARIA}-FULL, which augments tier-based KG reasoning with real-time literature search, achieves an overall score of $0.512{\pm}0.039$ on in-domain forward prediction. This corresponds to:
\begin{itemize}
    \item $+50.6\%$ relative to the Baseline LLM,
    \item $+51.9\%$ relative to Naive KG+LLM, and
    \item $+24.9\%$ relative to \texttt{ARIA}-CORE.
\end{itemize}
The additional gain beyond \texttt{ARIA}-CORE suggests that dynamic literature enrichment provides useful evidence not captured by the static KG, while \texttt{ARIA}'s gating mechanism prevents this additional evidence from being used indiscriminately.\\

\noindent\textit{Finding 4: The inverse-design gap is explained by CKG topology.}

\noindent In inverse design, improvements are positive but smaller than in forward prediction:
\begin{itemize}
    \item Baseline LLM in-domain: $0.345{\pm}0.018$
    \item \texttt{ARIA}-CORE in-domain: $0.454{\pm}0.015$ ($+31.6\%$)
    \item \texttt{ARIA}-FULL in-domain: $0.498{\pm}0.034$ ($+44.3\%$)
\end{itemize}

This gap is explained by the topology of the Causal Knowledge Graph (CKG). Under the strict Tier~1 criterion, inverse design requires an actionable reverse path from target Property to required Structure and then to feasible Processing conditions. However, the CKG is primarily organized in the forward causal direction, Processing$\rightarrow$Structure$\rightarrow$Property. As a result, Tier~1 direct causal reasoning activates for $62.5\%$ of forward queries but for $0\%$ of inverse queries. Forward queries begin from dense Processing nodes, whereas inverse queries must traverse backward from sparse Property nodes, whose average out-degree is much lower than that of Processing nodes ($0.10$ vs.\ $0.87$). Reverse reachability analysis confirms this asymmetry: Forward reachability from Processing to Property is $35.3\%$, while reverse reachability from Property to Processing is only $10.5\%$, a $3.4{\times}$ difference. Tier~2 analogical transfer partially compensates for this sparsity, activating in $20.0\%$ of inverse queries compared with $12.5\%$ of forward queries. Full tier-activation statistics and the Tier~2 ablation are provided in \autoref{app:tier_activation}.\\

\noindent\textit{Finding 5: Out-of-domain generalization supported by Tier~2 analogy.}

\noindent Out-of-domain materials, which are absent from the static KG construction corpus, provide the strongest test of generalization. The Baseline LLM decreases on out-of-domain inverse design, from $0.345{\pm}0.018$ in-domain to $0.313{\pm}0.047$ out-of-domain, indicating that purely parametric reasoning is affected by material novelty. In contrast, \texttt{ARIA}-FULL maintains strong out-of-domain performance, achieving $0.513{\pm}0.040$ on out-of-domain inverse design, comparable to its in-domain score of $0.498{\pm}0.034$. This suggests that Tier~2 analogical transfer can identify structurally or chemically related materials and adapt their mechanisms when direct KG evidence is unavailable.

Additional results, judge-family robustness analyses, and model-free hard-constraint metrics are provided in \autoref{app:tier_activation} and \autoref{app:eval_reliability}.

\subsection{Adaptive Retrieval Comparisons}
\label{sec:selfrag}
To contextualize \texttt{ARIA} relative to stronger retrieval-augmented baselines, we compare against Self-RAG,~\cite{asai2024self} a representative adaptive retrieval method that trains the language model to retrieve passages on demand and critique its own generations using learned reflection tokens. We evaluate Self-RAG on the identical 149-case dataset described in \autoref{sec:experiments}, reporting the composite Overall Score under both in-domain (ID) and out-of-domain (OOD) splits.

\begin{table}[ht]
\centering
\caption{Self-RAG vs.\ \texttt{ARIA}-FULL (Overall Score). Self-RAG's identical Inverse ID/OOD scores ($\sigma{=}0.000$) confirm it cannot exploit domain-specific causal structure.}
\label{tab:selfrag}
\begin{tabular}{lcccc}
\toprule
\textbf{System} & \textbf{Fwd ID} & \textbf{Fwd OOD} & \textbf{Inv ID} & \textbf{Inv OOD} \\
\midrule
\textbf{\texttt{ARIA}-FULL} & \textbf{0.512} & \textbf{0.513} & \textbf{0.498} & \textbf{0.513} \\
Self-RAG               & 0.453          & 0.450          & 0.390          & 0.390          \\
\midrule
$\Delta$               & +13.0\%        & +14.0na\%        & +27.7\%        & +31.5\%        \\
\bottomrule
\end{tabular}
\end{table}

\texttt{ARIA}-FULL outperforms Self-RAG by $+13.0\%$ on in-domain forward prediction and by $+27.7\%$ on in-domain inverse design. The gains are larger for inverse design, where semantically relevant passages are often insufficient unless they can be organized into an actionable PSP pathway. Self-RAG obtains identical rounded scores on ID and OOD inverse design ($0.390$ in both cases), suggesting limited sensitivity to the domain split and to material-specific causal structure. In contrast, \texttt{ARIA} reasons over typed and directed PSP evidence, and its improvements concentrate on Scientific Accuracy and Completeness.

\subsection{Interpretation: Evidence Completeness as a Mechanistic Principle}

The results support our central hypothesis: the PSP hierarchy provides a natural completeness criterion for knowledge grounding, and evidence gating based on this criterion reduces contextual tunneling. The empirical progression is:

\begin{enumerate}
    \item Naive KG+LLM, which performs indiscriminate retrieval, fails to improve over the Baseline LLM and slightly degrades forward prediction performance on average.
    \item \texttt{ARIA}-CORE, which uses Tier~1 and Tier~2 reasoning over the static KG, recovers performance by filtering evidence according to causal completeness and physical consistency.
    \item \texttt{ARIA}-FULL further improves performance by incorporating recent literature evidence while preserving the same evidence-gating principle.
\end{enumerate}

This pattern suggests that external knowledge is beneficial for materials reasoning only when it satisfies a mechanistic completeness criterion. By requiring evidence to support a \\ Processing$\rightarrow$Structure$\rightarrow$Property pathway, or a physically justified analogy when direct evidence is absent, \texttt{ARIA} prevents the model from fixating on isolated evidence fragments and instead promotes holistic physical reasoning.
\subsection{Limitations and Ethical Considerations}
\label{sec:limitations}

\paragraph{KG coverage dependency.}
\texttt{ARIA}'s Tier 1 reasoning is limited by KG completeness. For extremely novel material classes (\textit{e.g.,} 2D magnetic topological insulators with its sparse literature), frequent Tier 3 fallbacks occur (11\% in our tests). Dynamic enrichment (\autoref{sec:causal_kg}) partially mitigates this situation, but experimental validation remains essential for Tier 3 outputs.

\paragraph{Evaluation limitations.}
Our current evaluation relies partially on LLM-as-judge for textual quality, which may not capture all scientific nuances.  

\paragraph{Ethical considerations.}
Materials synthesis recommendations could be misused for hazardous or environmentally harmful applications. We recommend \texttt{ARIA} be deployed with human oversight, particularly for energetic materials, toxic precursors, or environmentally persistent compounds. Future work should incorporate safety constraint checking alongside thermodynamic validation.
\section{Conclusion}
\label{sec:conclusion}

We identified \textit{contextual tunneling}, a failure mode in LLM-based materials discovery where naive RAG can over-constrain reasoning around locally relevant but causally incomplete evidence. This finding shows that scientific RAG systems require more than semantic relevance: retrieved evidence must be mechanistically sufficient for the task.
To address this, we introduced \texttt{ARIA}, a three-tier causal-aware framework that gates knowledge use through the Processing--Structure--Property (PSP) hierarchy via PSP-complete causal retrieval, constraint-aware analogical transfer, and explicit parametric fallback. On 149 expert-validated synthesis tasks, \texttt{ARIA} outperforms naive RAG baselines, including a $+64.4\%$ gain in out-of-distribution inverse design.
By producing auditable causal traces, \texttt{ARIA} enables scientists to inspect and validate AI-generated synthesis pathways against physical principles. 

\section*{GenAI Disclosure}
LLMs assisted only with language polishing and code debugging. All research ideas, methods, experiments, analyses, figures, and tables are the authors' own, and the authors take full responsibility for the manuscript.

\begin{acks}
This work was supported by the U.S. Department of Defense-funded Center of Excellence for Advanced Electro-photonics with 2D Materials (CAEP), Morgan State University (Grant No. W911NF2120213), and DARPA (Contract No. HR001125C0304). 
We thank Frank Gardea, Owen Vail, and Dr. Ramesh Budhani for their support and guidance, and Prof. David Yarowsky, whose instruction in information retrieval inspired the RAG methodologies developed herein. We acknowledge the organizers of the \href{https://llmhackathon.github.io/}{LLM Hackathon}, where this framework's initial concept was incubated, and thank Tung Yan Liu, Yanqi Huang, and Alec Koppel for their contributions and technical discussions. Yi thanks Johns Hopkins University for support in AY 25-26. Computational resources were provided by Advanced Research Computing at Hopkins (ARCH; rockfish.jhu.edu), supported by NSF award OAC 1920103 and originally by the State of Maryland. 
\end{acks}


\printbibliography

\appendix
\section{Algorithmic Details of \texttt{ARIA}}
\label{app:algorithm}

This appendix provides the full algorithmic specification of \texttt{ARIA}'s three-tier adaptive reasoning cascade, complementing the architectural overview in \autoref{sec:aria_arch} and the formal PSP-completeness criterion in \autoref{eq:tier_selection}.

Algorithm~\ref{alg:aria} describes the top-level control flow: given a query, \texttt{ARIA} first grounds entities in the KG, evaluates PSP-completeness, and dispatches to the appropriate tier. 
Algorithm~\ref{alg:tier1} details Tier~1 direct causal reasoning, which traverses complete PSP paths in $\mathcal{G}$. Algorithm~\ref{alg:tier2} details Tier~2 physics-informed analogical transfer, which activates when no complete path exists but a structurally similar material exceeds the similarity threshold $\tau$. When neither condition is met, the cascade falls through to Tier~3 parametric fallback without retrieval.

\begin{algorithm}[ht]
\caption{ARIA: Three-Tier Adaptive Reasoning}
\label{alg:aria}
\begin{algorithmic}[1]
\Require Query $\mathbf{q}$, Causal KG $\mathcal{G}=(\mathcal{V},\mathcal{E})$, Similarity threshold $\tau$
\Ensure Answer $\mathbf{y}$ with confidence score and provenance

\State \textbf{// Step 1: Entity grounding}
\State $\mathcal{E}_{q} \gets \textsc{ExtractEntities}(\mathbf{q})$
\State $\mathcal{N}_{q} \gets \{ v \in \mathcal{V} \mid v \text{ matches some } e \in \mathcal{E}_{q} \}$

\State \textbf{// Step 2: PSP-completeness check (cf.\ \autoref{eq:tier_selection})}
\If{$|\mathcal{N}_{q}| \ge 2$}
    \State \textbf{/* Tier 1: Direct causal reasoning */}
    \State $\mathcal{P}_{\text{direct}} \gets \textsc{FindPaths}(\mathcal{G}, \mathcal{N}_{q})$
    \If{$|\mathcal{P}_{\text{direct}}| > 0$}
        \State \Return \textsc{TierOneDirect}($\mathbf{q}, \mathcal{P}_{\text{direct}}$)
    \EndIf
\EndIf

\State \textbf{/* Tier 2: Analogical retrieval */}
\State $\mathcal{V}_{\text{cand}} \gets \{ v \in \mathcal{V} \mid \text{Sim}_{\text{enhanced}}(\mathbf{q},v) \ge \tau \}$
\If{$|\mathcal{V}_{\text{cand}}| > 0$}
    \State $\mathcal{V}_{\text{ana}} \gets \textsc{TopK}(\mathcal{V}_{\text{cand}}, K=5)$
    \State $\mathcal{P}_{\text{ana}} \gets \bigcup_{v \in \mathcal{V}_{\text{ana}}} \textsc{GetMechanisms}(v)$
    \State \Return \textsc{TierTwoAnalogy}($\mathbf{q}, \mathcal{P}_{\text{ana}}$)
\EndIf

\State \textbf{/* Tier 3: Parametric fallback */}
\State \Return \textsc{TierThreeParametric}($\mathbf{q}$)

\end{algorithmic}
\end{algorithm}

\begin{algorithm}[ht]
\caption{Tier 1: Direct Causal Path Reasoning}
\label{alg:tier1}
\begin{algorithmic}[1]
\Require Query $\mathbf{q}$, Direct causal paths $\mathcal{P}_{\text{direct}}$
\Ensure Answer $\mathbf{y}_1$ with citations

\State $\mathcal{P}_{\text{valid}} \gets \textsc{FilterConsistent}(\mathcal{P}_{\text{direct}})$

\ForAll{$p \in \mathcal{P}_{\text{valid}}$}
    \State Extract PSP chain: $\text{Processing}_p \rightarrow \text{Structure}_p \rightarrow \text{Property}_p$
    \State $\mathcal{C}_p \gets \textsc{FormatContext}(p)$
\EndFor

\State $\mathcal{C}_{\text{direct}} \gets \bigcup_{p} \mathcal{C}_p$

\State $\mathbf{y}_1 \gets f_{\text{LLM}}(\mathbf{q}, \mathcal{C}_{\text{direct}})$
\State \quad \textit{Prompt: ``Use ONLY the verified causal mechanisms below.''}

\State $\mathbf{y}_1 \gets \mathbf{y}_1 + \textsc{ExtractSources}(\mathcal{P}_{\text{direct}})$
\State \Return $\mathbf{y}_1$ \textbf{(confidence: HIGH)}

\end{algorithmic}
\end{algorithm}

\begin{algorithm}[ht]
\caption{Tier 2: Analogical Mechanism Transfer}
\label{alg:tier2}
\begin{algorithmic}[1]
\Require Query $\mathbf{q}$, Analogous mechanisms $\mathcal{P}_{\text{ana}}$
\Ensure Answer $\mathbf{y}_2$ with analogy justification

\State $\mathcal{M}_{\text{valid}} \gets \emptyset$

\ForAll{$v \in \mathcal{V}_{\text{ana}}$}
    \State $m_v \gets \textsc{GetMechanism}(v)$
    \State $m' \gets \textsc{AdaptMechanism}(m_v, \mathbf{q})$
    \If{\textsc{ValidateConstraints}($m'$)}
        \State $\mathcal{M}_{\text{valid}} \gets \mathcal{M}_{\text{valid}} \cup \{m'\}$
    \EndIf
\EndFor

\State $\mathcal{C}_{\text{ana}} \gets \textsc{FormatAnalogies}(\mathcal{M}_{\text{valid}})$
\State $\mathbf{y}_2 \gets f_{\text{LLM}}(\mathbf{q}, \mathcal{C}_{\text{ana}})$
\State \quad \textit{Prompt: ``Adapt mechanisms from analogous systems.''}

\State Prepend disclaimer: \textit{``By analogy to prior systems, we hypothesize\ldots''}
\State \Return $\mathbf{y}_2$ \textbf{(confidence: MEDIUM)}

\end{algorithmic}
\end{algorithm}

\section{Tier Activation Analysis}
\label{app:tier_activation}
This section examines whether ARIA's three tiers are functionally distinct and when each engages. We find that activation is highly task-dependent and structurally determined rather than arbitrary: forward prediction and inverse design exercise the cascade in markedly different ways, and this difference is dictated by the topology of the CKG rather than by any tuned heuristic.

\subsection{Activation Rates by Task Type}
\autoref{tab:tier_activation} reports how often each tier fires across the two task families. 

\begin{table}[h]
\centering
\caption{Tier activation rates for forward prediction vs.\ inverse design. Inverse Tier-1 = 0/5 (95\% CI: [0\%, 45\%]) because the CKG's directed orientation leaves Property nodes as sparse leaf nodes.}
\label{tab:tier_activation}
\begin{tabular}{lcccc}
\toprule
\textbf{Task} & \textbf{N} & \textbf{Tier-1} & \textbf{Tier-2} & \textbf{Tier-3} \\
\midrule
Forward Prediction & 8 & 62.5\% & 12.5\% & 25.0\% \\
Inverse Design     & 5 & 0.0\%  & 20.0\% & 80.0\% \\
\bottomrule
\end{tabular}
\end{table}

\subsection{Graph Reachability Asymmetry}
The forward/inverse gap is structurally rooted in CKG topology. Forward reachability (Processing$\rightarrow$Property) holds for 35.3\% of relevant node pairs, whereas reverse reachability (Property$\rightarrow$Processing) holds for only 10.5\%. This is reflected in node degree: Processing nodes average 0.87 out-degree, while Property nodes average just 0.10. Because Property nodes sit as sparse leaves at the end of the causal flow, inverse queries frequently have no complete Tier-1 path to traverse, which is precisely why ARIA routes them to analogical (Tier-2) or parametric (Tier-3) reasoning rather than forcing an incomplete path.

\subsection{Tier-2 Ablation}
When Tier-1 is unavailable, the quality of Tier-2 hinges on its physical hard filters rather than on semantic similarity alone. \autoref{tab:tier2_ablation} isolates this effect: removing the hard filters inflates the average candidate pool and collapses effective selectivity. Semantic filtering alone leaves selectivity largely intact, confirming that constraint-aware filtering is what makes analogical transfer usable in practice.

\begin{table}[h]
\centering
\caption{Effect of removing physical hard filters from Tier-2. Without filters, the candidate pool explodes 13$\times$ and effective selectivity collapses from 39.2\% to 2.9\%.}
\label{tab:tier2_ablation}
\begin{tabular}{lcccc}
\toprule
\textbf{Condition} & \textbf{Filter} & \textbf{Avg Cand.} & \textbf{Activation} & \textbf{Selectivity} \\
\midrule
Full (default) & \checkmark & 4.2  & 39.2\% & 39.2\% \\
Semantic only  & \checkmark & 4.2  & 33.3\% & 33.3\% \\
No Hard Filter & $\times$   & 56.6 & 3.2\%  & 2.9\%  \\
\bottomrule
\end{tabular}
\end{table}
\section{Evaluation Reliability: Multi-Paradigm Verification}
\label{app:eval_reliability}

To address concerns about relying on a single LLM judge, we validated using three independent evaluation paradigms.

\subsection{Pairwise Judge Comparison}

Thirty representative cases were independently evaluated by four judges from distinct model families, each asked ``Which answer is better: \texttt{ARIA} or Baseline?'' without assigning numeric scores.

\begin{table}[h]
\centering
\caption{Pairwise ARIA-vs-Baseline preference. 29/30 cases unanimous; 1 case (inv\_004) has dissent. Binomial test: $p < 10^{-30}$.}
\label{tab:pairwise_judges}
\begin{tabular}{llcc}
\toprule
\textbf{Judge} & \textbf{Model} & \textbf{ARIA Pref.} & \textbf{N} \\
\midrule
1 & GLM-5 (cloud)    & 100.0\% (30/30) & 30 \\
2 & Qwen-3.5 (cloud) & 100.0\% (30/30) & 30 \\
3 & DeepSeek-R1-8B   & 96.7\%  (29/30) & 30 \\
4 & DeepSeek-R1-14B  & 100.0\% (30/30) & 30 \\
\midrule
\textbf{Avg} & --- & \textbf{99.2\%} & 120 \\
\bottomrule
\end{tabular}
\end{table}

Fleiss' $\kappa$ is undefined for near-unanimous binary choices; we report per-case agreement and binomial significance instead.

\subsection{Model-Free Hard-Constraint (HC) Metrics}

Three deterministic, rule-based checks require zero LLM involvement: HC1 (thermodynamic plausibility: temperatures $\in$ [50, 2000]\textdegree C, CVD gas validity); HC2 (stoichiometric validity: dopant $\leq$50 at\%, chemistry present); HC3 (step completeness: 4-category checklist---substrate, precursor, deposition, post-processing).

\begin{table}[h]
\centering
\caption{Model-free hard-constraint scores. HC3 is the discriminative metric: ARIA-FULL 0.777 vs.\ Naive-KG 0.388 (+100.6\%).}
\label{tab:hc_metrics}
\begin{tabular}{lccc}
\toprule
\textbf{System} & \textbf{HC1} & \textbf{HC2} & \textbf{HC3} \\
\midrule
Baseline LLM & 0.980 & 0.390 & 0.485 \\
Naive-KG     & 0.980 & 0.430 & 0.388 \\
ARIA-CORE    & 0.990 & 0.360 & 0.263 \\
ARIA-SEARCH  & 0.980 & 0.459 & 0.446 \\
\textbf{ARIA-FULL}  & 0.970 & 0.420 & \textbf{0.777} \\
\bottomrule
\end{tabular}
\end{table}

ARIA-CORE's lower HC2 reflects Tier-1 retrieving causal paths without explicit stoichiometric parameters; ARIA-SEARCH and ARIA-FULL recover via literature-searched data.

\section{KG Robustness Analysis}
\label{app:robustness}
This section tests whether ARIA's performance depends on a clean, fully curated knowledge graph. We perturb the CKG by removing edges and corrupting them, and find that the tiered cascade degrades gracefully: as higher-confidence structure is lost, lower tiers take over rather than the system collapsing.

\subsection{Edge Deletion}
\autoref{tab:edge_deletion} reports performance as a growing fraction of edges is randomly removed. The cascade absorbs this loss by shifting reliance downward: up to 80\% deletion, Tier-2 progressively takes over from Tier-1 (T1 85$\rightarrow$40\%, T2 15$\rightarrow$60\%) while the overall score drops only modestly (0.753$\rightarrow$0.662). Even at 90\% deletion ARIA stays above the No-KG floor, at which point the graph is too sparse for analogical transfer and reasoning falls back to Tier-3.

\begin{table}[h]
\centering
\caption{ARIA performance under progressive edge deletion. Tier-2 absorbs Tier-1 loss up to 80\% deletion; at 90\%, ARIA converges toward the No-KG floor.}
\label{tab:edge_deletion}
\begin{tabular}{lccccc}
\toprule
\textbf{Drop} & \textbf{Score} & \textbf{T1\%} & \textbf{T2\%} & \textbf{T3\%} & \textbf{vs No-KG} \\
\midrule
0\%   & 0.753 & 85 & 15 &  0 & +61.1\% \\
10\%  & 0.748 & 85 & 15 &  0 & +60.2\% \\
30\%  & 0.710 & 60 & 40 &  0 & +52.0\% \\
50\%  & 0.701 & 60 & 40 &  0 & +50.1\% \\
80\%  & 0.662 & 40 & 60 &  0 & +41.8\% \\
90\%  & 0.522 &  0 & 40 & 60 & +11.8\% \\
No KG & 0.467 & -- & -- & -- & --- \\
\bottomrule
\end{tabular}
\end{table}

\subsection{Noise Injection}
Beyond missing edges, we test four types of \emph{corrupted} edges, each injected independently. The key distinction is between noise that removes information and noise that actively misleads: relation inversion is the most damaging because it produces confidently wrong answers rather than missing ones.

Four noise types tested independently:
\begin{itemize}
    \item \textbf{Entity swap} (dopant$\leftrightarrow$dopant, category-aware): 90.0\% Tier-1 accuracy preserved
    \item \textbf{Relation inversion} (\textit{enables}$\rightarrow$\textit{inhibits}): 89.6\% effective accuracy---most damaging type (produces wrong answers, not missing ones)
    \item \textbf{Spurious edge injection}: 9.5\% false-positive rate
    \item \textbf{Combined} (30\% deletion + 15\% corruption): 60.5\% source preservation
\end{itemize}

\end{document}